\documentclass[letterpaper, 10 pt, conference]{ieeeconf}  

\usepackage{times}
\usepackage{textcomp}
\usepackage{amsmath}
\usepackage{amssymb}
\usepackage{graphicx}
\usepackage[pagebackref=true,breaklinks=true,letterpaper=true,colorlinks,bookmarks=false]{hyperref}
\usepackage{booktabs}     
\usepackage{enumerate}    
\usepackage{cite}         
\usepackage{tikz}
\usepackage{fix2col}
\usepackage{pgfplots}
\usepackage{subcaption}

\usetikzlibrary{arrows,positioning} 
\usetikzlibrary{calc,patterns,decorations.pathmorphing,decorations.markings,fit,backgrounds}
\tikzset{
    >=stealth',
    box/.style={
           rectangle,
           rounded corners,
           draw=black, very thick,
                          minimum height=1cm,
                          minimum width=1.5cm,
           text centered},
    state/.style={
                rectangle,
                rounded corners,
                draw=black, very thick,
                minimum height=0.6cm,
                minimum width=1.0cm,
                text centered},
    pil/.style={
           ->,
           thick,
           shorten <=2pt,
           shorten >=6pt,}

}

\IEEEoverridecommandlockouts                              

\title{\LARGE \bf A Flexible Modeling Approach for Robust Multi-Lane Road Estimation}

\author{Alexey Abramov$^{\S}$, Christopher Bayer$^{\S}$, Claudio Heller$^{\S}$ and Claudia Loy$^{\S}$
\thanks{${\S}$ These authors contributed equally to this work.}
\thanks{Alexey Abramov, Christopher Bayer, Claudio Heller and Claudia Loy are with Continental Teves AG, Chassis \& Safety Division, Advanced Engineering,
	Guerickestrasse 7, DE-60488, Frankfurt am Main, Germany. \tt\small{\{alexey.abramov, christopher.bayer, claudio.heller, claudia.loy\} @continental-corporation.com}}
}

\begin{document}

\maketitle
\thispagestyle{empty}
\pagestyle{empty}

\begin{abstract}
A robust estimation of road course and traffic lanes is an essential part of environment perception for next generations of Advanced Driver Assistance Systems and development of self-driving vehicles.
In this paper, a flexible method for modeling multiple lanes in a vehicle in real time is presented.
Information about traffic lanes, derived by cameras and other environmental sensors, that is represented as features, serves as input for an iterative expectation-maximization method to estimate a lane model.
The generic and modular concept of the approach allows to freely choose the mathematical functions for the geometrical description of lanes.
In addition to the current measurement data, the previously estimated result as well as additional constraints to reflect parallelism and continuity of traffic lanes, are considered in the optimization process.
As evaluation of the lane estimation method, its performance is showcased using cubic splines for the geometric representation of lanes in simulated scenarios and measurements recorded using a development vehicle.
In a comparison to ground truth data, robustness and precision of the lanes estimated up to a distance of 120~m are demonstrated.
As a part of the environmental modeling, the presented method can be utilized for longitudinal and lateral control of autonomous vehicles.
\end{abstract}

\section{INTRODUCTION AND RELATED WORK \label{introduction}}
Advanced Driver Assistance Systems in modern vehicles are aimed to help drivers and increase comfort and safety for them and their passengers. Systems like lane departure warning, lane keeping assist, adaptive cruise control, emergency brake assist and blind spot monitoring are designed to support drivers in keeping their car within the lane and avoiding collisions with other traffic participants and static objects~\cite{Bengler2014}. The next generation of driver assistance technologies and autonomous vehicles require comprehensive and extensive knowledge about the environment. One of its fundamental parts is a continuous and robust perception of the road and its lanes. This includes the reliable representation of multiple lanes on various types of roads with high detection range and availability.


Lane estimation approaches consist of several processing parts: detection of lane markings based on sensors, possible fusion and accumulation of lane information, followed by modeling of lanes. The latter presumes building a mathematical model which describes course of the road and relevant traffic lanes.
What kind of lane geometries can be represented accurately, depends on the choice of the lane model.
A variety of models have been utilized for lane modeling in the past: straight lines, parabolic curves, clothoids and splines.
The simple straight~\cite{Lakshmanan1996} and parabolic~\cite{McCall2006} lane models are computationally efficient and very robust against measurement noise, but they only can model certain road shapes and do not have sufficient flexibility for representing a wider range of lane geometries. Clothoids are traditionally used for highway scenarios, since they are utilized in the design and construction of this type of roads~\cite{Dickmanns1992,Meis2010,TeamMuc2016}. Despite a high precision on ordinary highway sections, a single clothoid model cannot represent highways in full detail. For instance, it is often not flexible enough to model construction zones, junctions, entrance and exit ramps. Furthermore, on rural roads the clothoid model sometimes fails or cannot be set up with desired distance and accuracy.


A common way to model curves is using piecewise defined functions such as splines, which are composed of connected polynomials. They have proven to have the capacity to approximate complex road shapes with high curvatures, e.g. double bends or sharp turns. Splines are more flexible but also more sensitive to uncertainties in measurement data. Some state-of-the-art approaches model only the ego lane~\cite{Wang2004,Kim2008}. Others can handle roads with multiple lanes~\cite{Aly2008,Huang2009}, but do not consider semantic information, such as lane marking type, and geometric correlation between lanes are not considered.

In the focus of this paper is a generic approach for robust and continuous modeling of multiple traffic lanes on highways and rural roads. The method builds on features describing lane markings which can be provided by common lane detection techniques. Centerpiece is an iterative lane model estimation method which allows for a free choice of mathematical model. Beyond position and orientation of the measured features, the modeling procedure incorporates supplementary information, such as parallelism, tracking and continuity of lanes. By means of the presented approach, lane estimation can be performed online in a vehicle without requiring prior knowledge about course of the road or number of existing traffic lanes. The outcome of the system can be used for localization, trajectory planning and control of self-driving vehicles. Evaluation of the modeling method is performed in simulated scenarios and on real sensor data, where the modeling result is compared to ground truth.


A description of techniques for estimating the lane model can be found in section \ref{lane-model-estimation}, whereas the evaluation methods and results are presented in section \ref{evaluation}, followed by the conclusion in section \ref{conclusion}.

\section{LANE MODEL ESTIMATION\label{lane-model-estimation}}
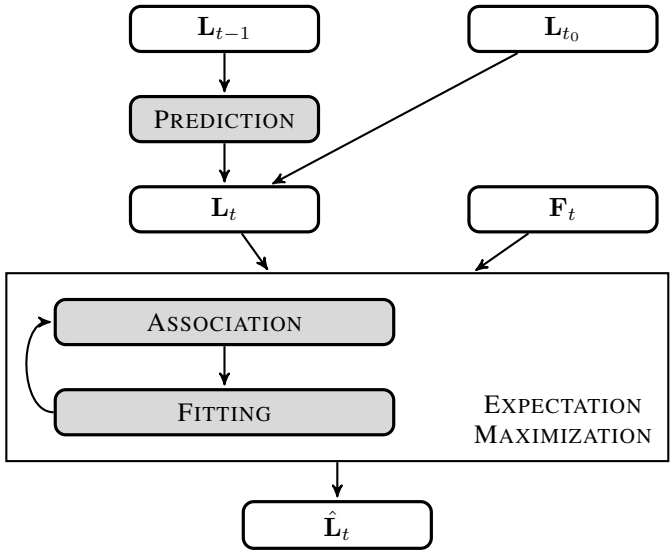
\begin{figure}
\begin{tikzpicture}[->,>=stealth',shorten >=1pt,auto,node distance=3cm,
  thick]

  \node[state] 
  	(PrevLaneModel) 
  	[minimum width=2.5cm] 
  	{$\mathbf{L}_{t-1}$};

  \node[state] 
  	(Prediction) 
  	[below of=PrevLaneModel,  node distance=1.2 cm, minimum width=2.5cm, fill =  black!15!white]
  	{\textsc{Prediction}}; 
  
  \node[state] 
  	(LaneModel) 
  	[below of=Prediction,  node distance=1.2 cm, minimum width=2.5cm]
  	{$\mathbf{L}_{t}$};
  	
  \node[state] 
  	(InitLaneModel) 
  	[right of=PrevLaneModel,  node distance=4.5 cm, minimum width=2.5cm]
  	{$\mathbf{L}_{t_0}$};  	
  	
  \node[state] 
  	(InputData) 
    [right of=LaneModel,  node distance=4.5 cm, minimum width=2.5cm]
    {$\mathbf{F}_t$};  	 	
  	
 \node[draw=black,rectangle, minimum width=8.8cm, minimum height=2.5cm] 
 	(EMBox) at (1.5,-4.5) {}; 		
 	
 \node[below of = InputData,  node distance=2.6 cm] 
 	(EstimationText) {\textsc{Expectation}}; 	
 \node[below of = EstimationText,  node distance=0.4 cm] 
 	(MaximizationText) {\textsc{Maximization}}; 
  	
  	
  \node[state] 
  	(Association) 
  	[below of = LaneModel,  node distance=1.5 cm, minimum width=4.5cm, fill =  black!15!white]
  	{\textsc{Association}}; 

  	
  \node[state] 
  	(Fitting) 
  	[below of = Association,  node distance=1.2 cm, minimum width=4.5cm, fill =  black!15!white]
  	{\textsc{Fitting}}; 

  	
  \node[state] 
  	(FinalLaneModel) 
  	[below of = EMBox,  node distance=2.1 cm, minimum width=2.5cm]
  	{$\hat{\mathbf{L}}_t$};


  \path[every node/.style={font=\sffamily\small}]
  	(PrevLaneModel) 	edge[black] node [] {} (Prediction)
  	(Prediction) 		edge[black] node [] {} (LaneModel)
    (InitLaneModel) 	edge[black] node [] {} (LaneModel)
    (LaneModel)		 	edge[black] node [] {} (EMBox)
    (InputData) 		edge[black] node [] {} (EMBox)
    (Association) 		edge[black] node [] {} (Fitting)
    (Fitting) 		    edge[black, bend left = 90] node [] {} (Association)
    (EMBox) 		    edge[black] node [] {} (FinalLaneModel)
    ;

    
\end{tikzpicture}
\centering
\caption{Architectural overview of the iterative lane modeling method. 
Either the previous model $\mathbf{L}_{t-1}$ is predicted to the current time step 
or an initial model $\mathbf{L}_{t_0}$ is derived from the current input data $\mathbf{F}_t$. After that the expectation maximization algorithm performs the \textit{Association}
and \textit{Fitting} iteratively until the final solution $\hat{\mathbf{L}}_t$ is estimated.
}

\label{fig:architecture}
\end{figure}
The goal of the method presented in this study is to obtain a robust representation of traffic lanes on a road that can be used for trajectory planning, longitudinal and lateral control of a vehicle and localization within a map. The resulting lane model (see section \ref{lane-model-describtion}) is composed of several mathematical functions each depicting a lane marking which separates two lanes or delimits a lane. Thus, the lane model contains information about number, position and geometry of detected lanes with respect to the vehicle as well as attributes of lane markings, such as type (dashed, solid, block) and color.
To find the optimal model, its parameters are estimated in an iterative expectation-maximization (EM) process \cite{dempster1977maximum} that alternates between associating the input data (expectation) to the current model and fitting a new model to the associated data (maximization).

\begin{figure}
		\centering
		\begin{subfigure}{0.5\textwidth}
			\includegraphics[width=8.5cm]{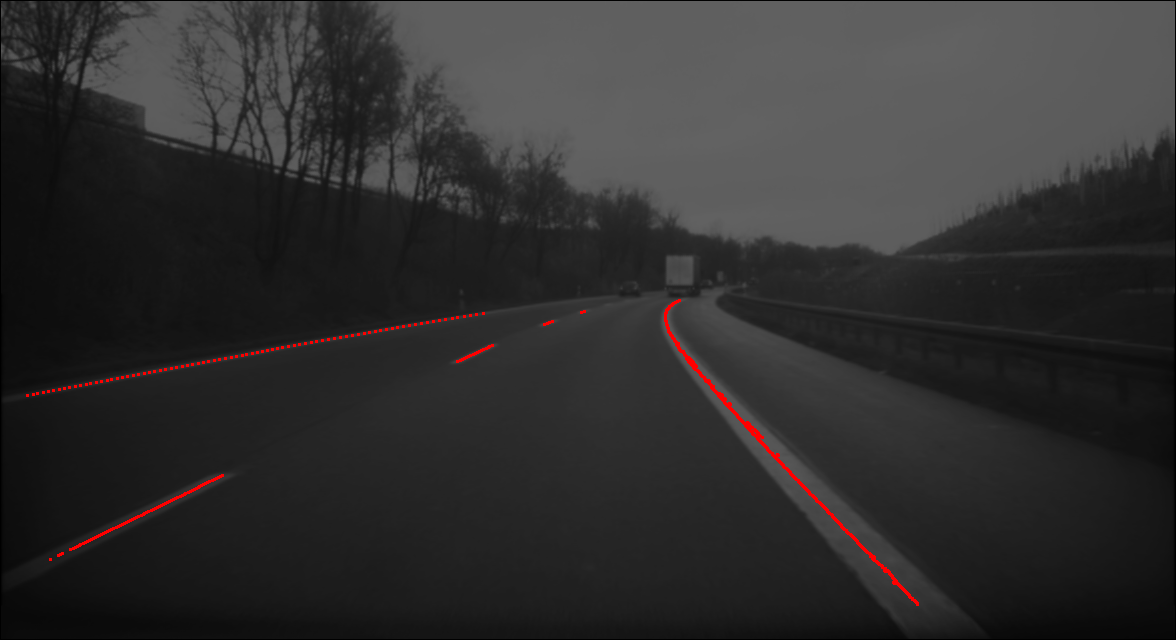}
			\centering
			\caption{Features (red) extracted on the lane markings in a camera image.}
			\label{fig:feature_image}
			\vspace{0.2cm}
		\end{subfigure}
		\begin{subfigure}{0.5\textwidth}
			\newlength\figureheight 
			\newlength\figurewidth 
			\setlength\figureheight{3cm} 
			\setlength\figurewidth{7cm}
			\input{hdmap_17_s195.tikz}
			\centering
			\caption{Lane modeling result (black) based on the extracted features (red) in vehicle coordinates.}
			\label{fig:feature_modeling}
	    \end{subfigure}
	    \caption{Examplary snapshot of feature extraction and optimized lane model.}
	    \label{fig:exemplary_modeling}
\end{figure}

An architectural overview of the lane modeling method is shown in fig. \ref{fig:architecture}. 
At the beginning of each processing loop, a lane model is needed for association of the input data $\mathbf{F}_t$. If a model $\mathbf{L}_{t-1}$ was derived at the previous time step, it serves as initial model after predicting it to the current time step as described in section \ref{model-prediction}. If no previous model is available, an initial model $\mathbf{L}_{t_0}$ is estimated based on the present input data (see section~\ref{lane-model-init}). The input data $\mathbf{F}_t$ is comprised of a vector of lane features that describe the lane markings and which are defined in section \ref{input-data}. During the expectation step the lane features are associated to the lane model. As described in detail in section \ref{gauss-newton-fit}, in the maximization part a new lane model is estimated taking into account the associated input data and the result from the previous time step.
This process is repeated iteratively until there is no change in the association of the input data. The optimized result $\hat{\mathbf{L}}_t$ for the current time step corresponds to the last estimated lane model.
Fig. \ref{fig:exemplary_modeling} shows an example of features extracted in a camera image (fig. \ref{fig:feature_image}) and the resulting optimized lane model (fig. \ref{fig:feature_modeling}).

\subsection{Input Data \label{input-data}}
The input data for the lane model estimation provides information about lane markings on the road that the vehicle is currently driving on. This information can, for instance, be obtained by a camera system and image processing methods. In the scope of this work, lane information is derived by fusion of data from various input sensors as described in \cite{TeamMuc2016}. Another possibility is the utilization of information provided by maps as (additional) input for the modeling. To be independent from the source and type of the lane information, the interface of the input data is defined in a general way. It is specified as a feature vector $\mathbf{F}_t$, which contains lane features
\begin{equation}
\mathbf{f}_i = [x_i,y_i,\theta_i,\mathbf{\Sigma}_i, \mathbf{a}_i ] \in \mathbf{F}_t,
\label{def.feature}
\end{equation}
where $x_i$ and $y_i$ constitute the position of a feature $\mathbf{f}_{i}$ and $\theta_i$ constitutes its heading in the vehicle coordinate system~\footnote{The vehicle coordinate system $(x,y,z)$ is a right-handed coordinate system, its origin lies in the middle of the front axle (height of the road), $x$ is identical to the driving direction, $y$ points to the left and $z$ points upwards.}. The measurement uncertainty of each feature is given by a covariance matrix $\mathbf{\Sigma}_i \in \mathbb{R}^{3\times3}$ with respect to $x$, $y$ and $\theta$. Additional information such as color of a feature or the corresponding lane marking type is described by the attributes $\mathbf{a}_i$. Note that this two-dimensional feature representation of lanes is free of geometric model assumptions.

Additionally to the lane features, the odometry of the vehicle is utilized in the process of lane estimation.

\subsection{Lane Model Description \label{lane-model-describtion}} 
\begin{figure}[t]
	\begin{center}
		\includegraphics[width=8.7cm]{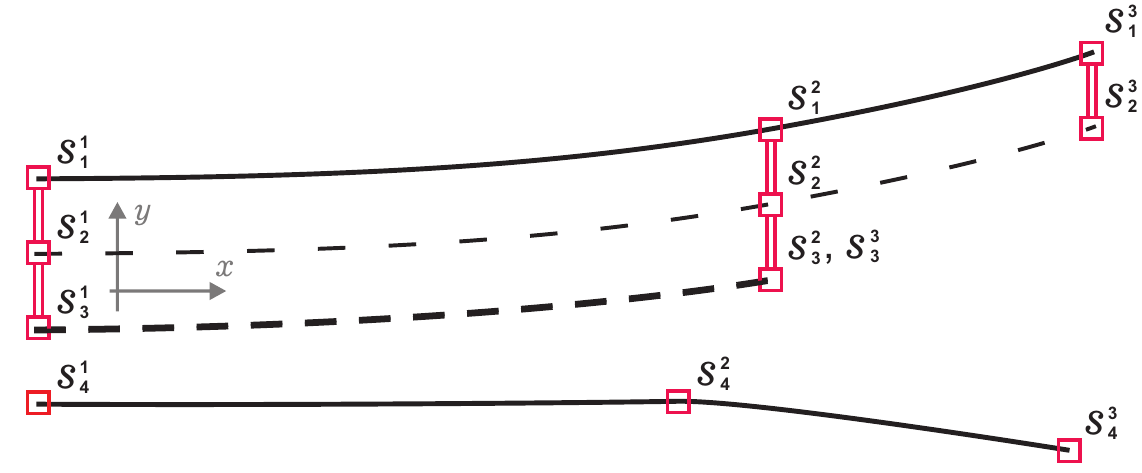}
		\caption{Example of a lane model with two driving lanes and an exit ramp. This model consists of four lines ($N=4$) with two segments ($M=2$) and three control points $s_{n}^m$ each. Note that $s_{3}^3$ equals $s_{3}^2$ due to the length of that line. The marking type of the lines are either solid ($n=1$, $n=4$), dashed ($n=2$) or block ($n=3$). The vertical red double lines (e.g. from $s_{1}^1$ to $s_{2}^1$) indicate parallelism between the connected segments. The vehicle coordinate system is denoted in gray.}
		\label{fig:RoadModelExample}
	\end{center}
\end{figure}

For the modeling of lanes a generic representation is used and an example is shown in fig. \ref{fig:RoadModelExample}. 
The lane model consists of $n=1...N$ lines each composed of $m=1...M$ segments. 
Each segment is described by a mathematical function $f(x,\mathbf{l}^m_n)$, which depends on the parameters $\mathbf{l}^m_n$ and is computed in $x$.
Therefore, the lane model $\mathbf{L}_t$ is defined by the stacked parameter vector $[\mathbf{l}^1_1,\mathbf{l}^2_1,...,\mathbf{l}^M_1,...,\mathbf{l}^1_N,\mathbf{l}^2_N,...,\mathbf{l}^M_N]^T$.
Within the scope of this work each line is represented as a cubic spline. Thus, third order polynomials are used as functions to describe the segments
\[ 
f(x,\mathbf{l}^m_n) = a x^3 + b x^2 + c x + d,~x \in [ s^{m-1}_n, s^m_n [~, 
\]
where $\mathbf{l}^m_n = [a,b,c,d]^T$ are the parameters of the function. 
The functions of two subsequent segments are continuous at the $M+1$ control points $s^m_n$ with respect to position, heading and curvature. 
Additionally, each line has attributes for type (solid, dashed and block~\footnote{In some countries entry and exit lanes are separated from usual driving lanes by wide dashed markings called block markings}) and color to describe the corresponding lane marking properties.

\subsection{Initial Lane Model \label{lane-model-init}}
An essential part of the lane estimation method, is the association between the present lane feature input and the current best estimate of the lane model (see section \ref{model-feature-association}). As no previous estimate is available in the first algorithm loop, an initial model needs to be derived. Therefore, the input features in close proximity to the vehicle are projected onto the lateral axis of the vehicle coordinate system to obtain several separated distributions. If enough features contribute to such a distribution, mean lateral position and orientation of the features are computed. Assuming that roads are rather straight within a short distance, each line in the lane model is initialized as a straight line at the computed lateral offset with the corresponding heading. Additionally, this method is applied at every time step to all features that could not be associated to the previously estimated lane model for the case that new lane markings accrue.

\subsection{Prediction \label{model-prediction}}
Due to the movement of the vehicle between two subsequent time steps, it is necessary to transform the lane model, obtained in the previous time step, to the current vehicle coordinate system. 
The prediction of $\mathbf{L}_{t-1}$ is conducted based on the change in odometry and yields a lane model $\mathbf{L}_t$ that can be used for association of the input data $\mathbf{F}_t$. In order to keep the mathematical form of the lane representation when rotating to another coordinate system, an approximate prediction is performed. For a lane model composed of cubic splines, each spline segment is predicted by transforming its limiting control points to the current vehicle coordinate system and performing a fit to position and heading of the transformed points. Note that a cubic spline transformed using this approximation is not continuous in curvature anymore, but it is still sufficiently precise for association of the input data.

\subsection{Association and Model Assumptions \label{model-feature-association}}
In the expectation part of the EM method, the association is performed by determining the correspondences, which maximize the likelihood given the current lane model estimate $\mathbf{L}_{t}$ and the current lane features $\mathbf{F}_t$:
\[ 
\hat{\mathbf{c}}_t = \underset{\mathbf{c}_t}{\arg\max} \: p(\mathbf{c}_t |\mathbf{F}_t,\mathbf{L}_{t}).
\]
The correspondence vector $\mathbf{c}_t$ consists of tuples $\left\langle \mathbf{f}_i, \mathbf{l}_n^m  \right\rangle$ of a lane feature and the associated line segment and can be estimated by minimizing the distance between those.
Therefore, the Mahalanobis distance~\cite{Mahalanobis1936} is evaluated between a feature and a line with respect to position and orientation. Furthermore, to reject outliers, an upper limit in the Euclidean distance of $2~m$ is used as additional criteria~\footnote{Traffic lanes usually have a width between $3.5~m$ to $4~m$.}. If no association can be found for enough lane features, an attempt to set up a new line of the lane model using the method described in section \ref{lane-model-init} is made and, if successful, the association procedure is repeated. To prevent limiting the association to the range of the previous lane model, each line is extrapolated in longitudinal direction to also associate lane features that lie beyond that range. This means that the scope of the lane model can be increased with every EM iteration. 

After association, the current range of each lane model line is determined and its attributes are derived based on the associated lane features. The range of a line is given by the associated features with the shortest and largest longitudinal distance. Lane marking type and color of the lines are determined fusing the corresponding information of the associated features according to Dempster-Shafer theory \cite{Shafer1976}.

Based on the derived lane marking type, assumptions about parallelism between adjacent lines in the lane model are made. From a dashed narrow line one can conclude that it is parallel to its left and right adjacent lines. A continuous line on the other hand indicates lanes that might separate into different directions and therefore no parallelism of lines beyond the continuous one is assumed. In the case of block markings, also no parallelism is assumed beyond the corresponding line. An example of grouping the lines of a lane model according to parallelism based on the lane marking type is shown in fig.~\ref{fig:RoadModelExample}, where the upper three lines are parallel. Inferred information about parallel lines enters the lane model fit as equality constraint (see section \ref{gauss-newton-fit}).

After this, the positions of control points connecting segments of a line are determined. In the case of parallel lines, a control point is set at the end of each line. 
In fig.~\ref{fig:RoadModelExample} the longitudinal position of $s_1^2$, $s_2^2$ and $s_3^2$ are equal and correspond to the end of the block marking. To provide flexibility to the lane model, the length of line segments is restricted by usage of additional control points where necessary (e.g. $s_4^2$ in fig.~\ref{fig:RoadModelExample}).

\subsection{Lane Model Fit Using Constrained Gauss Newton Method \label{gauss-newton-fit}}
The goal of the maximization in the EM algorithm is to find an optimal model to the input data.
In general this task can be formulated as maximizing the probability distribution
\begin{equation}
\hat{\mathbf{L}}_{t} = \underset{\mathbf{L}_{t}}{\arg \max} \: p(\mathbf{L}_{t}|\mathbf{F}_t, \mathbf{c}_t),
\label{eq:maximization}
\end{equation}
which yields the optimized lane model $\hat{\mathbf{L}}_{t}$, given the input features $\mathbf{F}_t$ and the association $\mathbf{c}_t$ between input features and lines of the current model. 
In the following, a description for incorporating different information and constraints into the optimization problem is given.
\subsubsection{Measurement optimization}
To find the lane model which represents the current lane features in the best way, eq. \ref{eq:maximization} is formulated as a quadratic minimization problem
\begin{equation}
\hat{\mathbf{L}}_{t} = \underset{\mathbf{L}_{t}}{\arg \min} 
\sum_{\left\langle \mathbf{f}_i, \mathbf{l}_n^m \right\rangle \in \mathbf{c}_t} \mathbf{e}(\mathbf{f}_i,\mathbf{l}_n^m)^T \mathbf{\Omega}_{i} \mathbf{e}(\mathbf{f}_i,\mathbf{l}_n^m),
\label{eq:minimization}
\end{equation}
where $\mathbf{c}_t$ is the correspondence vector determined in the expectation step (see section \ref{model-feature-association}).
The error function 
\[
\mathbf{e}(\mathbf{f}_i,\mathbf{l}_n^m) = 
\left[ 
\begin{array}{c}
f(x_i,\mathbf{l}_n^m) - y_i \\ 
f'(x_i,\mathbf{l}_n^m) - \theta_i
\end{array} 
\right]
\]
is defined as the distance of position and heading between a lane feature ($x_i,y_i,\theta_i \in \mathbf{f}_i$) and the associated line segment.
It is multiplied from both sides to the information matrix $\mathbf{\Omega}_{i} = diag([\sigma_{y}^2, \sigma_{\theta}^2])^{-1}$ which corresponds to the variances of feature $\mathbf{f}_i$.

The sum over the non-linear quadratic equations in eq. \ref{eq:minimization} can be solved by the Gauss-Newton algorithm as shown for example for pose graph optimization \cite{Grisetti2010}. Given an initial guess $\breve{\mathbf{L}}_t$, the solution to the minimization problem can be found iteratively by solving
\begin{equation}
\mathbf{H}
\Delta \mathbf{L}^{\ast}_t = -\mathbf{b}
\label{eq:solve_system}
\end{equation}
with 
\begin{equation}
\mathbf{H} = \sum_{\left\langle \mathbf{f}_i, \breve{\mathbf{l}}_n^m   \right\rangle \in \mathbf{c}_t} \mathbf{J}(\mathbf{f}_i, \breve{\mathbf{l}}_n^m)^T \mathbf{\Omega}_{i} \mathbf{J}(\mathbf{f}_i,\breve{\mathbf{l}}_n^m)
\label{eq:H-matrix}
\end{equation}
and
\begin{equation}
\mathbf{b} = \sum_{\left\langle \mathbf{f}_i , \breve{\mathbf{l}}_n^m \right\rangle \in \mathbf{c}_t} \mathbf{e}(\mathbf{f}_i, \breve{\mathbf{l}}_n^m)^T \mathbf{\Omega}_{i} \mathbf{J}(\mathbf{f}_i, \breve{\mathbf{l}}_n^m).
\label{eq:b-vector}
\end{equation}
Here $\mathbf{J}(\mathbf{f}_i, \breve{\mathbf{l}}_n^m)$ is the Jacobian of the error function $\mathbf{e}(\mathbf{f}_i,\breve{\mathbf{l}}_n^m)$ evaluated at the current estimate $\breve{\mathbf{L}}_t$. 
After solving eq.~\ref{eq:solve_system} for $\Delta \mathbf{L}^{\ast}_t$, the current estimate is updated
\[ 
\mathbf{L}^{\ast}_t = \breve{\mathbf{L}}_t + \Delta \mathbf{L}^{\ast}_t
\]
and used in the next iteration as initial guess.
After convergence (no change in the parameter update $\Delta \mathbf{L}^{\ast}_t$), the optimized solution $\hat{\mathbf{L}}_t = \mathbf{L}^{\ast}_t$ is found.
\subsubsection{Time filter optimization}
In addition, the previous state $\mathbf{L}_{t-1}$ is considered in the optimization to ensure continuity of the result over time. 
As shown in section \ref{model-prediction} the control points of the previously derived lane model can be predicted to the current time step. 
Therefore, an additional sum of error terms
\[
\sum_{n,m} \mathbf{e}(\check{\mathbf{s}}_n^m, \mathbf{l}_n^m)^T \mathbf{\Omega}_n^m \mathbf{e}(\check{\mathbf{s}}_n^m,\mathbf{l}_n^m)
\]
is added to the optimization (eq. \ref{eq:minimization}), where $\check{\mathbf{s}}_n^m$ corresponds to position and orientation of the predicted control point. 
The information matrix $\mathbf{\Omega}_n^m$ regarding the lateral displacement and orientation of the control point is calculated by the inverse of the previous system matrix $\mathbf{H}_{t-1}$ (eq. \ref{eq:H-matrix}), the Jacobian of the line function and the Jacobian of the coordinate transformation.
Incorporation of these error terms into eq. \ref{eq:solve_system} prevents large jumps in position and heading of the lane model between two subsequent time steps.

\subsubsection{Continuity and Parallelism as equality constraints}
In addition to previously formulated optimization problem, equality constraints can be added to limit the state space of the lane model.
On the one hand the function defining the geometry of a line needs to be continuous in position, heading and curvature at the control points. 
On the other hand lanes on highways are often parallel and therefore this parallelism needs to be taken into account while solving the optimization problem. 

As described in section \ref{lane-model-describtion}, the lane model is composed of $N$ lines with $M$ segments each. 
For each pair of successive line segments $\left\langle m,m+1 \right\rangle$ the continuity with respect to position, orientation and curvature is established by the equality constraints 
\[ 
f(s^{m+1}_n,\mathbf{l}^m_n) - f(s^{m+1}_n,\mathbf{l}^{m+1}_n) \overset{!}{=} 0
\]
\[ 
f'(s^{m+1}_n,\mathbf{l}^m_n) - f'(s^{m+1}_n,\mathbf{l}^{m+1}_n) \overset{!}{=} 0
\]
\begin{equation}
f''(s^{m+1}_n,\mathbf{l}^m_n) - f''(s^{m+1}_n,\mathbf{l}^{m+1}_n) \overset{!}{=} 0,
\label{eq:continuity_constraint}
\end{equation}
where $s^{m+1}_n$ is the longitudinal position of the control point between the two segments.
\\
\\
For arbitrary line functions 
the equality constraints can be incorporated by extending the system of eq.~\ref{eq:solve_system} using the method of Lagrange multipliers \cite{griva2009linear}.
In general, the quadratic function from eq. \ref{eq:minimization} subject to the constraints $\mathbf{g}(\mathbf{L}_t) = \mathbf{0}$ 
can be iteratively minimized by solving the linear equation system
\begin{equation} 
\left[ 
\begin{array}{cc}
\mathbf{H} & -\mathbf{K}^T \\ 
-\mathbf{K} & \mathbf{0}
\end{array}  
\right] 
\left[ 
\begin{array}{c}
\Delta \mathbf{L}_t^\ast\\ 
\boldsymbol{\lambda}
\end{array} 
\right]
= 
\left[
\begin{array}{c}
-\mathbf{b} \\ 
\mathbf{g}(\breve{\mathbf{L}}_t)
\end{array} 
\right],  
\label{eq:constraint_GN}
\end{equation}
for the state space update $\Delta \mathbf{L}_t^\ast$ and $\boldsymbol{\lambda}$ as the Lagrange multiplier.
$\mathbf{K}$ is the Jacobian of $\mathbf{g}(\breve{\mathbf{L}}_t)$ evaluated at the current state estimate $\breve{\mathbf{L}}_t$. $\mathbf{H}$ and $\mathbf{b}$ are the matrices defined for the measurement optimization in eq.~\ref{eq:H-matrix} and eq.~\ref{eq:b-vector}. Using this method to integrate the equality constraints of eq.~\ref{eq:continuity_constraint} into the optimization problem,  $\mathbf{g}(\mathbf{x})$ is a vector of $3N(M-1)$ equations.

For certain functions, the linear equation system resulting from eq.~\ref{eq:continuity_constraint} can be solved explicitly.
For a cubic polynomial the eight parameters 
($[\mathbf{l}^{m}_n, \mathbf{l}^{m+1}_n]^T = [a_n^m, b_n^m, c_n^m, d_n^m, 
a_n^{m+1}, b_n^{m+1}, c_n^{m+1}, d_n^{m+1}]^T$)
of two successive segments can be reduced to five applying the substitutions:
\[ 
c_n^{m+1} = c_n^m + 3 (d_n^m - d_n^{m+1}) s^{m+1}_n,
\]
\[ 
b_n^{m+1} = b_n^m + 2(c_n^m - c_n^{m+1}) s^{m+1}_n + 3(d_n^m - d_n^{m+1}) (s^{m+1}_n)^2,
\]
\[ 
a_n^{m+1} = a_n^m + (b_n^m - b_n^{m+1}) s^{m+1}_n  + (c_n^m - c_n^{m+1})(s^{m+1}_n)^2
\]
\[ 
+ (d_n^m  - d_n^{m+1}) (s^{m+1}_n)^3.
\]
This method has two advantages. 
First, the state space of $\mathbf{L}_t$ in eq. \ref{eq:constraint_GN} is reduced from $4MN$ parameters to $4 M N - 3(M-1)N =  (M+3)N$. Additionally in this case the $3(M-1)N$ equations for $\mathbf{g}(\mathbf{L}_t)$ are not needed, which makes the solving of the linear system computationally faster.
\\
To also incorporate parallelism constraints between neighboring lines, the vector of equality equations $\mathbf{g}(\mathbf{L}_t)$ can be extended.
For cubic splines the degree of freedom of the model can be considered to find the number of necessary constraints.
Two cubic spline lines have $8M$ individual parameters, respectively $2(M+3)$ parameters after including continuity constraints with the substitutions.
If these lines should be parallel, the degree of freedom needs to be reduced to two lateral offsets for left and right line, heading, curvature and curvature derivate for the first segment and one curvature change per following segment: 
\[  
\left|  [d^1_n, d^1_{n+1}, c^1, d^1, a^1, a^2, ... , a^{m}]^T \right|  = (M+4).
\]
Therefore, $2(M+3) - (M+4) = M+2$ constraints need to be added to the system of eq. \ref{eq:constraint_GN}, three for the first segment and one per additional segment, where a constraint is defined as the equality of the orientation:
\[ 
g(\mathbf{l}_i, \mathbf{l}_j) = 
f'(x, \mathbf{l}_i) - f'(x, \mathbf{l}_j) 
\overset{!}{=} 0.
\]
The evaluation point $x$ corresponds either to the control points or in the first segment also to the middle of the two limiting control points.
Note that the parallelism criteria is approximated by demanding equality of orientation.

\section{EXPERIMENTAL EVALUATION \label{evaluation}}
Several tests have been performed to evaluate the presented lane modeling method.
In the first section, the modeling of a simulated double bend is analyzed comparing two different modeling functions.
In the second section, the performance of the lane detection system is evaluated using sensor data collected on German highways~\footnote{A video sequence illustrating the results: \url{https://drive.google.com/open?id=0B8eccpkD0x9nSGJEZExhTmZhUm8}.}.

\subsection{Lane Modeling on Simulated Data \label{simulation-eval}}

\begin{figure}[t]
	\centering
	\setlength\figureheight{5cm} 
	\setlength\figurewidth{7cm}
	\input{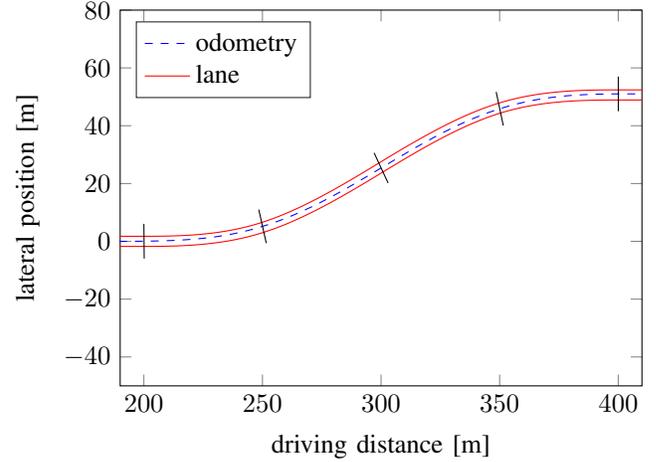}
	\caption{Double bend of the simulated scenario. The double bend consists of four connected  $50~m$ long clothoid segments. Along each segment the radius of curvature changes from $1000~m$ to $100~m$ or vice versa.}
	\label{fig:simulated-scenario}
\end{figure}

\begin{figure}[t]
	\centering
	\setlength\figureheight{5cm} 
	\setlength\figurewidth{7cm}
	\input{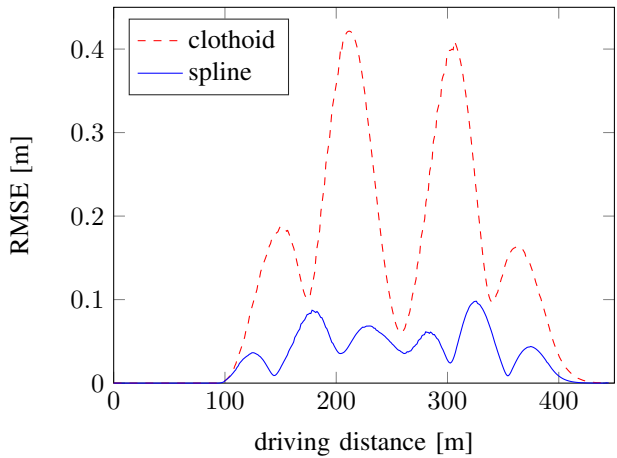}
	\caption{RMSE of the difference between modeled and simulated lanes. The result is shown for clothoids (red) and cubic splines (blue) in a simulated scenario containing a double bend.}
	\label{fig:rmse-sim}
\end{figure}

To demonstrate the functioning and the flexibility of the presented method, a spline and a clothoid \footnote{Here, clothoids are approximated by a third order polynomial~\cite{Dickmanns1992}.} model are compared using a simulated scenario.
Due to the generic description of the lane model, clothoids can be modeled with the proposed method by simply using a third order polynomial as line function with one segment.
The CarMaker simulation software~\footnote{http://ipg.de/de/simulation-software/carmaker/} is used to generate odometry and lane features of the simulated scenario. It consists of a $200~m$ long straight road section followed by a double bend, which is composed of four connected clothoid segments. The s-shape curve of the simulated scenario is shown in fig. \ref{fig:simulated-scenario}. At each time step, lane features up to a longitudinal distance of $100~m$ in front of the vehicle serve, together with the current odometry, as input for the lane modeling. As a measure of performance, the Root Mean Square Error (RMSE) of the lateral distance between the estimated lane model and simulated lane features is computed. Fig. \ref{fig:rmse-sim} shows the measured RMSE for the resulting lanes when using clothoids and cubic splines in the modeling method. As expected, there is no difference between the two models for the straight road section. However, one can clearly see the benefit of the spline model once the curve comes into range. The maximum error for the clothoid model is approximately four times as high as the one for the spline model. Note that the cubic spline model is not able to perfectly describe the simulated scenario, as its control points are not positioned at the connection points of the simulated clothoids. Nevertheless, using the spline model the lanes are modeled at any point of the simulated scenario with an RMSE below $0.1~m$.



%


\subsection{Lane Estimation Evaluation on Sensor Data \label{real-data-eval}}

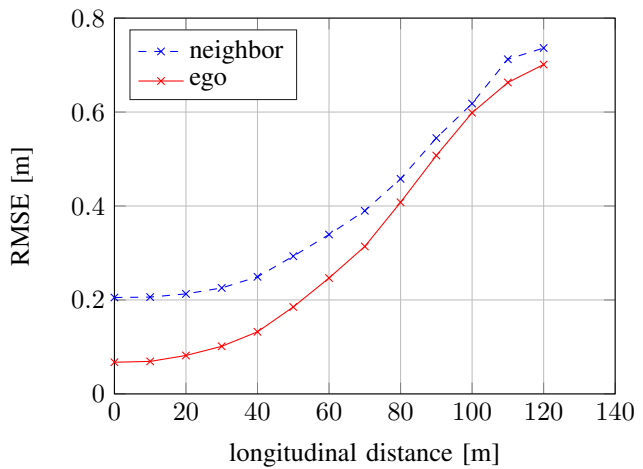
\begin{figure}[t]
	\centering
	\setlength\figureheight{5cm} 
	\setlength\figurewidth{7cm}
%
%
\begin{tikzpicture}

\begin{axis}[%
width=0.951\figurewidth,
height=\figureheight,
at={(0\figurewidth,0\figureheight)},
scale only axis,
xmin=0,
xmax=140,
xlabel={longitudinal distance [m]},
xmajorgrids,
ymin=0,
ymax=0.8,
ylabel={RMSE [m]},
ymajorgrids,
axis background/.style={fill=white},
legend style={at={(0.03,0.97)},anchor=north west,legend cell align=left,align=left,draw=white!15!black}
]
\addplot [color=blue,dashed,mark=x,mark options={solid}]
  table[row sep=crcr]{%
0.000119999999999898	0.205192346611507\\
10.00012	0.206112184969043\\
20.00012	0.212943104639537\\
30.00012	0.22559510727648\\
40.00012	0.249256961718107\\
50.00012	0.293165195757767\\
60.00012	0.339235374556829\\
70.00012	0.389830123534918\\
80.00012	0.457820923269442\\
90.00012	0.544645440891243\\
100.00012	0.6181410269616\\
110.00012	0.712363912206831\\
120.00012	0.736331750788874\\
};
\addlegendentry{neighbor};

\addplot [color=red,solid,mark=x,mark options={solid}]
  table[row sep=crcr]{%
0.000119999999999898	0.0674000517973819\\
10.00012	0.0692201992472808\\
20.00012	0.0817956496691816\\
30.00012	0.101311883784134\\
40.00012	0.132157736443648\\
50.00012	0.185021209944948\\
60.00012	0.246651975460887\\
70.00012	0.31389483807362\\
80.00012	0.407892915604124\\
90.00012	0.50742778074888\\
100.00012	0.59855935761095\\
110.00012	0.662900407464778\\
120.00012	0.701182319098093\\
};
\addlegendentry{ego};

\end{axis}
\end{tikzpicture}%
	\caption{RMSE of the difference between estimated lanes and ground truth for ego (red) and neighbor lane (blue).}
	\label{fig:rmse-real}
\end{figure}

In the analysis described in the following, the performance of lane feature fusion \cite{TeamMuc2016} and subsequent lane modeling using the proposed modeling method with cubic splines is evaluated on real data measurements. The modeled lanes are compared to a ground truth map that contains global positions of lane markings as a point vector. The analyzed route is a highway in Germany with three lanes and left and right curves~\footnote{The analysis presented in \cite{TeamMuc2016} is based on the same data and ground truth map.}. The map is generated using a high precision GPS system, which is also used for localization in the map. The input data has been collected during several drives on the highway with a development vehicle corresponding to a total driving distance of $24~km$. The development vehicle is equipped with camera and radar systems to detect lane markings and other traffic participants. The sensor information is fused in a GraphSLAM based process that yields the lane features which serve as input for the modeling. In the analysis, the lateral deviation of the modeled lanes to the relevant map points is computed at each time step. The result is accumulated in dependence of the longitudinal distance for all of the recorded data. As a measure of performance the RMSE is determined within distance intervals of $10~m$. Fig. \ref{fig:rmse-real} shows the result for ego and adjacent lanes up to a longitudinal distance of $120~m$. Due to higher precision of the input data, the result for the ego lane is better in comparison to the one for adjacent lanes. At a distance of $120~m$ lanes are modeled with an RMSE of less than $0.75~m$. Compared to the result presented in \cite{TeamMuc2016}, the application of the proposed modeling approach using cubic splines provides a similar performance but offers a higher degree of flexibility as shown in section \ref{simulation-eval} and in the following.


\begin{figure}[t]
	\centering
	\begin{subfigure}{0.5\textwidth}
		\includegraphics[width=8.5cm]{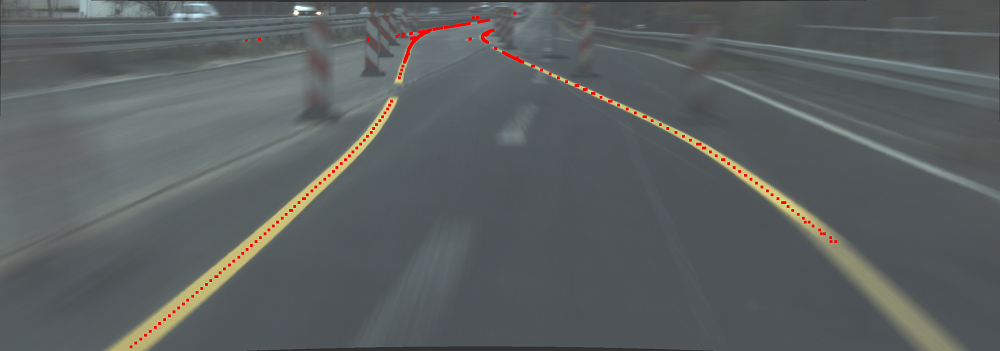}
		\centering
		\caption{Snapshot of lane features (red) detected in a camera image.}
		\label{fig:features_from_longrange}
		\vspace{0.2cm}
	\end{subfigure}
	\begin{subfigure}{0.5\textwidth}
		\setlength\figureheight{3cm} 
		\setlength\figurewidth{7cm}
		\input{lane_model_baustelle.tikz}
		\caption{Modeling of camera features accumulated over time (red).}
		\label{fig:modeling_construction_site}
	\end{subfigure}
	\caption{Example of lane modeling based on cubic splines inside a construction zone.}
	\label{fig:construction_site}
\end{figure}

In fig. \ref{fig:construction_site} an example of feature extraction and lane model estimation result inside a construction zone is shown. In this setting the lane markings inside the construction zone are yellow and have a double bend shape. Features are extracted along the lane markings in the camera image (fig.~\ref{fig:features_from_longrange}). After accumulation and fusion of the features, the lanes are modeled using the presented method with cubic splines and three segments (fig.~\ref{fig:modeling_construction_site}). As one can see, the resulting lane model is properly describing the shape of the lane within the construction zone. Despite outliers in the input data, a robust estimation of the lane is obtained.

\section{CONCLUSION \label{conclusion}}
In this work, a flexible real-time modeling method for robust estimation of ego and adjacent lanes is presented. An iterative expectation-maximization method is applied, which alternately associates the input data to the current model and estimates a new model by solving the corresponding constrained optimization problem. The underlying lane model is defined in a generic way as a composition of arbitrary mathematical functions. In the scope of this study, cubic splines are utilized in the approach to model traffic lanes. Evaluation of the method is shown in simulated scenarios as well as real data measurements that were collected with a development vehicle. In the latter, performance of the modeling method is analyzed by comparison of the result to ground truth data.

The results show that the method is capable of modeling multiple lanes on highways that include entry and exit lanes, transitions between roads and lanes with double bends, like construction sites. The precision and robustness achieved in modeling lanes up to a range of $120~m$ on highways suffice to be used in the development and testing of self-driving vehicles. In the recent months, the presented method in combination with a lane feature fusion algorithm has been applied to drive several thousand kilometers autonomously on highways.

Enhancement of the presented modeling method could be achieved by incorporating additional information to the optimization, such as constraining the lanes within detected road boundaries. To further improve robustness of the modeled lanes, boundary conditions, such as limiting curvature of the modeled lanes could be considered. Using the road curvature from a map would be a possibility to include prior knowledge for improvement of the result. In addition, the quality of the derived lane model depends on the input data and would therefore draw benefit from improved lane detection methods.




{\small
\bibliographystyle{IEEEtran}
\bibliography{iv2017_references}
}

\end{document}